\documentclass[letterpaper]{article} 
\usepackage{aaai25}  
\usepackage{times}  
\usepackage{helvet}  
\usepackage{courier}  
\usepackage[hyphens]{url}  
\usepackage{graphicx} 
\usepackage{natbib}  
\usepackage{caption} 
\usepackage{algorithm}
\usepackage{algorithmic}
\usepackage{color}
\usepackage{mathtools, amsmath, amssymb}
\usepackage{booktabs}
\usepackage{multirow}
\usepackage{float}
\usepackage{marvosym}

\usepackage{newfloat}
\usepackage{listings}
\DeclareCaptionStyle{ruled}{labelfont=normalfont,labelsep=colon,strut=off} 
\floatstyle{ruled}
\newfloat{listing}{tb}{lst}{}
\floatname{listing}{Listing}
\title{Mamba-CAD: State Space Model For 3D Computer-Aided Design Generative Modeling}
\author{
    Xueyang Li,
    Yunzhong Lou,
    Yu Song,
    Xiangdong Zhou\textsuperscript{\Letter}
}
\affiliations{

    School of Computer Science, Fudan University, Shanghai, China\\
    xueyangli21@m.fudan.edu.cn, yzlou20@fudan.edu.cn, songy23@m.fudan.edu.cn, xdzhou@fudan.edu.cn
}

\usepackage{bibentry}
\begin{document}
\maketitle
\begin{abstract}
Computer-Aided Design (CAD) generative modeling has a strong and long-term application in the industry. Recently, the parametric CAD sequence as the design logic of an object has been widely mined by sequence models. However,  the industrial CAD models, especially in component objects, are fine-grained and complex, requiring a longer parametric CAD sequence to define. To address the problem, we introduce Mamba-CAD, a self-supervised generative modeling for complex CAD models in the industry, which can model on a longer parametric CAD sequence. Specifically, we first design an encoder-decoder framework based on a Mamba architecture and pair it with a CAD reconstruction task for pre-training to model the latent representation of CAD models; and then we utilize the learned representation to guide a generative adversarial network to produce the fake representation of CAD models, which would be finally recovered into parametric CAD sequences via the decoder of Mamba-CAD. To train Mamba-CAD, we further create a new dataset consisting of 77,078 CAD models with longer parametric CAD sequences. Comprehensive experiments are conducted to demonstrate the effectiveness of our model under various evaluation metrics, especially in the generation length of valid parametric CAD sequences. The code and dataset can be achieved from https://github.com/Sunny-Hack/Code-for-Mamba-CAD-AAAI-2025-.
\end{abstract}

%
\section{Introduction}\label{in}
Computer-Aided Design (CAD) generative modeling has garnered widespread attention in the industry, particularly in fields such as manufacturing, vehicle engineering, and architectural design. The design logic of each object in the world can be expressed by a sequence of parametric commands. For example, constructing a cylinder is defined with the radius to draw a complete circle which would be further extruded with the determined height. Typically, the complex design logic requires a longer sequence of parametric commands while a simpler object only requires a shorter parametric CAD sequence to define. In CAD fields, this kind of representation is called parametric CAD sequences, allowing designers to use CAD tools (e.g., AutoCAD and SolidWorks) to quickly edit the model by modifying their commands and parameters. It is undeniable that prototypes of various objects in the world almost starts with CAD designs. Therefore, it attracts much attention of generative modeling on CAD models recently~\cite{c:19, c:20, r:25}. 
The parametric CAD sequence, as a key point used in CAD tools to create 3D shapes, can be seen as a special discrete language consisting of commands and their parameters. Hence, some interesting efforts~\cite{c:1, c:2, c:23} leverage Transformer-based architecture~\cite{r:4} to learn the representation of parametric CAD sequences and achieve impressive generation results. However, the parametric CAD sequences generated by them are not long enough to satisfy the construction of complex 3D shapes. In practical applications, especially in the industrial manufacturing, the design logic of objects is relatively complex, which necessitates longer parametric CAD sequences to define. This motivates us to make a step forward, trying to design a generative model with the capability of producing a longer parametric CAD sequence. Inspired by recent Mamba~\cite{r:8}, a State Space Model (SSM) performing well in modeling long sequences, we further explore its capability for the parametric CAD sequence modeling. Note that we mainly discuss CAD models defined with Construct Solid Geometry (CSG) representation as it is conducive to be parameterized, while Boundary Representation (BRep) for CAD models is beyond the scope of this paper as it lacks construction history to recover how the designs are created by CAD commands.
\begin{figure}[t]
\centering
\includegraphics[width=0.99\columnwidth]{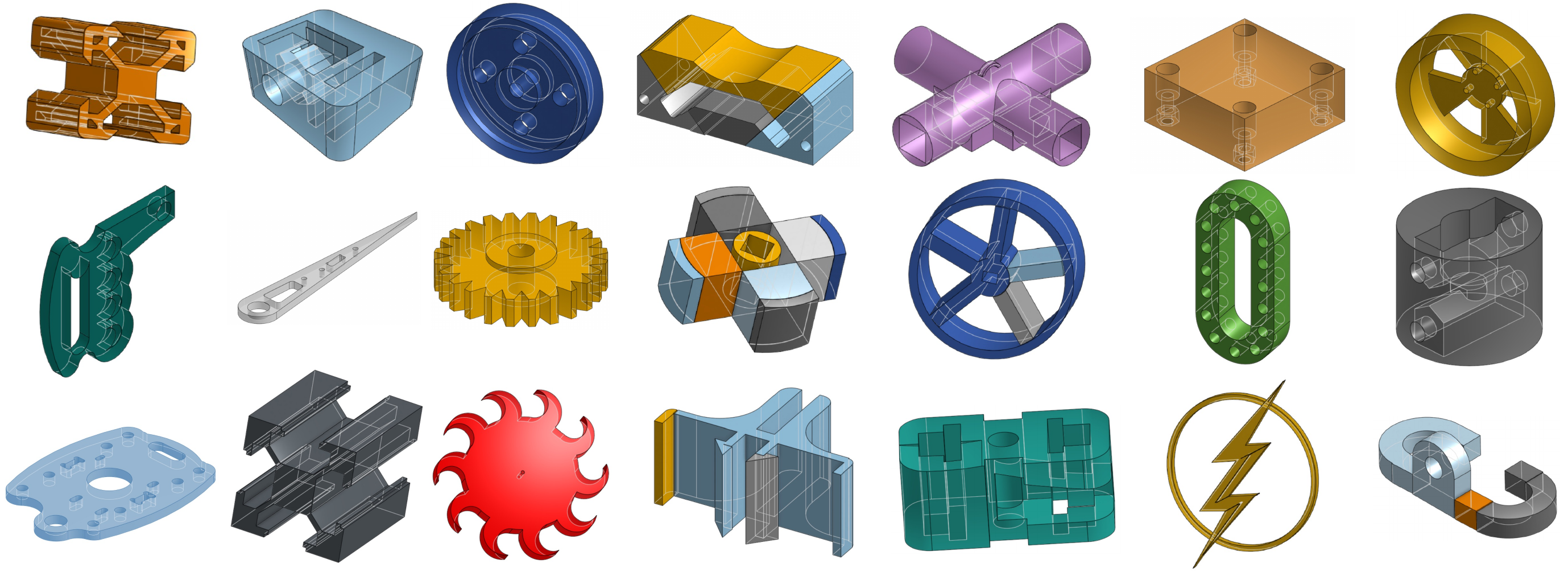} 
\caption{The gallery of CAD models created by Mamba-CAD, including visual illustrations of the created CAD models from CAD sequence reconstruction, completion, and random generation.}
\label{fig1}
\end{figure}

To overcome the challenge of generative modeling of CAD models, we propose Mamba-CAD, a self-supervised framework for a longer parametric CAD sequence generation, which is easy to implement and straightforward to train. Specifically, we first design an auto encoder-decoder architecture based on a Mamba backbone to process the parametric CAD sequence and pair it with a CAD reconstruction task for pre-training. When the network is well-trained, we further select the learned latent representation as a knowledgable distribution to train a latent generative adversarial network~\cite{c:16, c:17} to produce the fake representation of CAD models, which is randomly sampled from a standard gaussian distribution. Finally, these fake representations would be fed into the decoder of Mamba-CAD to generate parametric CAD sequences. Given that the benchmark DeepCAD dataset~\cite{c:1} only contains CAD models with the maximum 60 commands, it is hard to show the capability of Mamba-CAD in handling a longer parametric CAD sequence. Hence, we again filter the DeepCAD dataset by discarding some CAD models with a total number of commands less than 10, at the same time we use a feature script from~\cite{m:1} to parse CAD models from the ABC dataset~\cite{c:18} to retain samples with the total number of commands from 61 to 128. Finally, total 77,078 CAD models described as parametric CAD sequences are grouped into a new dataset that will be released publicly to promote future research on the parametric CAD modeling.

In summary, our main contribution is Mamba-CAD, a simple yet effective self-supervised framework for generative CAD modeling, which can handle a longer parametric CAD sequence to achieve more complex CAD models. To train Mamba-CAD, we further construct a new dataset of parametric CAD sequences that is more complex than the existing benchmark dataset of the same type and will make it publicly available in the future. Comprehensive experiments show that Mamba-CAD outperforms other competitive methods in tasks such as CAD sequence reconstruction, completion, and random generation, especially in the generation length of valid parametric CAD sequences.
\section{Related Work}
\subsection{Parametric CAD modeling}
 Boundary Representation (BRep) and Construct Solid Geometry (CSG) are two widely used representations in parametric CAD fields. For BRep representation, BRep-Bert~\cite{c:4} applies Graph Neural Network~\cite{r:1} as a tokenizer to process BRep representation and further enhance its feature representation via conducting the masked entity modeling. 
Complexgen~\cite{r:2} views the reconstruction process of CAD models as a chain complex in the BRep representation, achieving the task of CAD reconstruction from point cloud. Solidgen~\cite{r:3} employs Transformer~\cite{r:4} and Pointer network~\cite{r:5} to facilitate the generation of complex B-rep structures.  ~\cite{c:6} design a CADParser, a learning approach of sequence modeling for converting BRep data to CSG representation. For CSG representation,  CSG-Stump~\cite{c:7} simplifies the CSG representation into a three-layer structure consisting of union, intersection, and complement operation. SkexGen~\cite{c:2} utilizes two individual and distinct Transformer-based branches to learn representations of ``sketch" and ``extrude" operation separately. HNC-CAD~\cite{c:23} proposes a code tree representation learned by a two-stage cascaded auto-regressive networks. Besides, some concurrent works also explore reconstructing CAD models from point clouds, voxels, meshes, and images~\cite{c:8, c:9, r:6, c:10, r:7, c:24, c:25}. Unlike these efforts, our method mainly focuses on addressing the generation problem of a longer parametric CAD sequence based on CSG like representation. The most related to our work is DeepCAD~\cite{c:1}, a Transformer-based architecture, which encodes the parametric CAD sequence into a latent space and then decodes it back to the constructed 3D shape. Different from DeepCAD our framework is designed for generate longer parametric CAD sequence to construct a more complex 3D shape.
\subsection{Mamba-based models}
Mamba, a recent selective structured state space model~\cite{r:8} with excellent capabilities of processing long sequence data, has been widely explored in vision, language, and sequence modeling. For Mamba in the vision task, inspired by ViT~\cite{c:14}, VMamba~\cite{r:11} and ViMamba~\cite{r:9} are two early representative Mamba models that view several image patches as sequences from the horizontal and vertical dimensions of the image and perform bi-directional scanning on these two directions. Gradually, many interesting studies delve into the exploration of conducting Mamba backbone in image segmentation~\cite{r:10, r:14}, video understanding~\cite{r:12, r:13}, and 3D shape~\cite{r:20, r:21}. For language tasks, JMamba~\cite{r:15} consolidates Transformer layers, Mamba layers and a mixture-of-experts (MoE)~\cite{r:16, r:17} into a hybrid architecture. Similarly, several notable works make efforts in integrating or modifying components into Mamba backbone, achieving competitive performance across various language applications~\cite{c:15, r:18, r:19}. For sequence modeling, many interesting works leverages Mamba to capture long-term dependencies and patterns in time series data~\cite{r:22, r:23, r:24}, tacking the challenge of attention complexity in Transformer-based methods. Given Mamba is a sequence model with powerful capabilities of long sequence modeling, at the same time the parametric CAD model can be seen as a sequence consisting of series of commands and their parameters, which motivates us make a step forward to conduct Mamba in the task of generative CAD modeling.
\section{Method}
In this section, we now introduce our Mamba-CAD model. To learn Mamba-CAD capable of encoding and decoding CAD models, we adopt a well defined data structure namely parametric CAD sequences and make them discretized into friendly network representations, which will be fed forward into Mamba-CAD. To train our approach, we further create a new collection of 77,078 CAD models, in which the maximize length of parametric CAD sequences within the dataset is up to 128 and the average length of them is significantly longer than existing benchmark datasets of the same type, and we plan to make it publicly available shortly.
\subsection{CAD sequence representation}\label{sec2}
\begin{figure}[t]
\centering
\includegraphics[width=1.0\columnwidth]{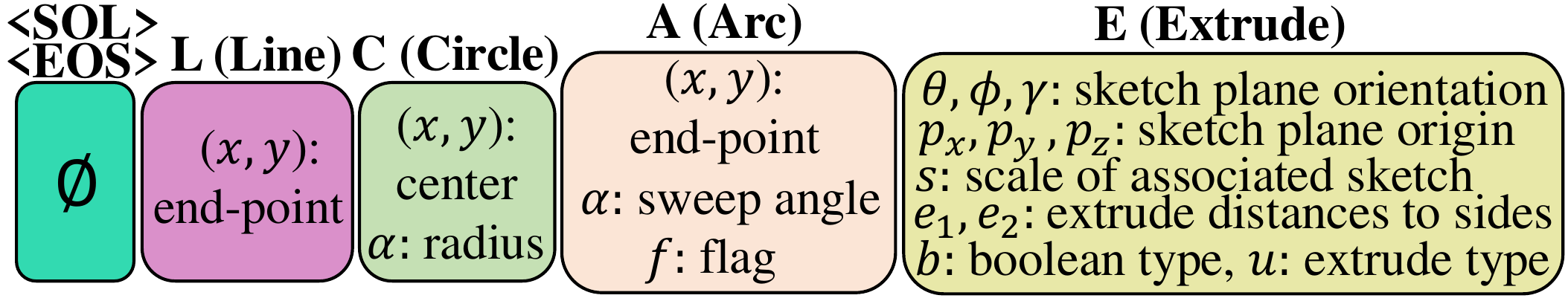} 
\caption{Common CAD commands discussed in this paper. $<$SOL$>$ and $<$EOS$>$ are two symbols for representing the start of one loop and the end of a whole parametric CAD sequence, respectively. $f$ is a counter-clockwise flag.}
\label{fig2}
\end{figure}

For each CAD model, its workflow construction can be expressed by a parametric CAD sequence as each object is constructed by iteratively adding features such as line, arc, extrusion, or other CAD operations. The parametric CAD sequence $S$ is kind of a discrete language consisting of a group of CAD commands $c_{i}\in C$ and their parameters $p_{i}\in P$, where $S=\left\{(c_{i}, p_{i})\in(C, P)|G[(c_{i}, p_{i})_{1}, (c_{i}, p_{i})_{2},\ldots,(c_{i}, p_{i})_{N}]\right\}$.The function of $G[*]$ is to gather $(c_{i}, p_{i})$ into a group in sequential order, thereby forming a complete parametric CAD sequence $S$. In the practical design, orderly executing the CAD commands $C$ defined in a parametric CAD sequence $S$ can construct the shape of each object. For the CSG representation discussed in this paper, the commonly used parametric CAD commands can be found in Figure~\ref{fig2}. Intuitively, parametric CAD sequence is kind of complex as its different commands are coupled with various parameters that are not in the same scale. For example, the command E owns 12 parameters while the command L only owns 2 parameters. Besides, the command E is composed of discrete  and continuous parameters simultaneously. These attributes increase the difficulty of using CAD commands directly for network learning. To make it friendly representation for deep learning networks, we fix the length of parametric CAD sequences up to $N=128$ and pad those short parametric CAD sequences to achieve the length of 128 via adopting the command $<$EOS$>$, the empty symbol for representing the end of a parametric CAD sequence. As seen in Figure~\ref{fig2}, the number of parameters varies for different types of CAD commands $c_{i}$. Again, for each CAD command, we fix the number of parameters into 16 including all parameters listed in Figure~\ref{fig2} and leverage scalar -1 to pad unused parameters. Each CAD command is finally stacked into a 16$\times$1 vector. To unify discrete and continuous values in parameters $p_{i}$, we refer DeepCAD~\cite{c:1} to quantize continuous values into a 256-dimensional vector using 8-bit integers after normalizing each CAD model into a $2\times2\times2$ cube. So far, the representation of each raw CAD model has been described in a series of discrete values.
\subsection{Mamba-CAD}\label{sec3}
\begin{figure*}[t]
\centering
\includegraphics[width=1.56\columnwidth]{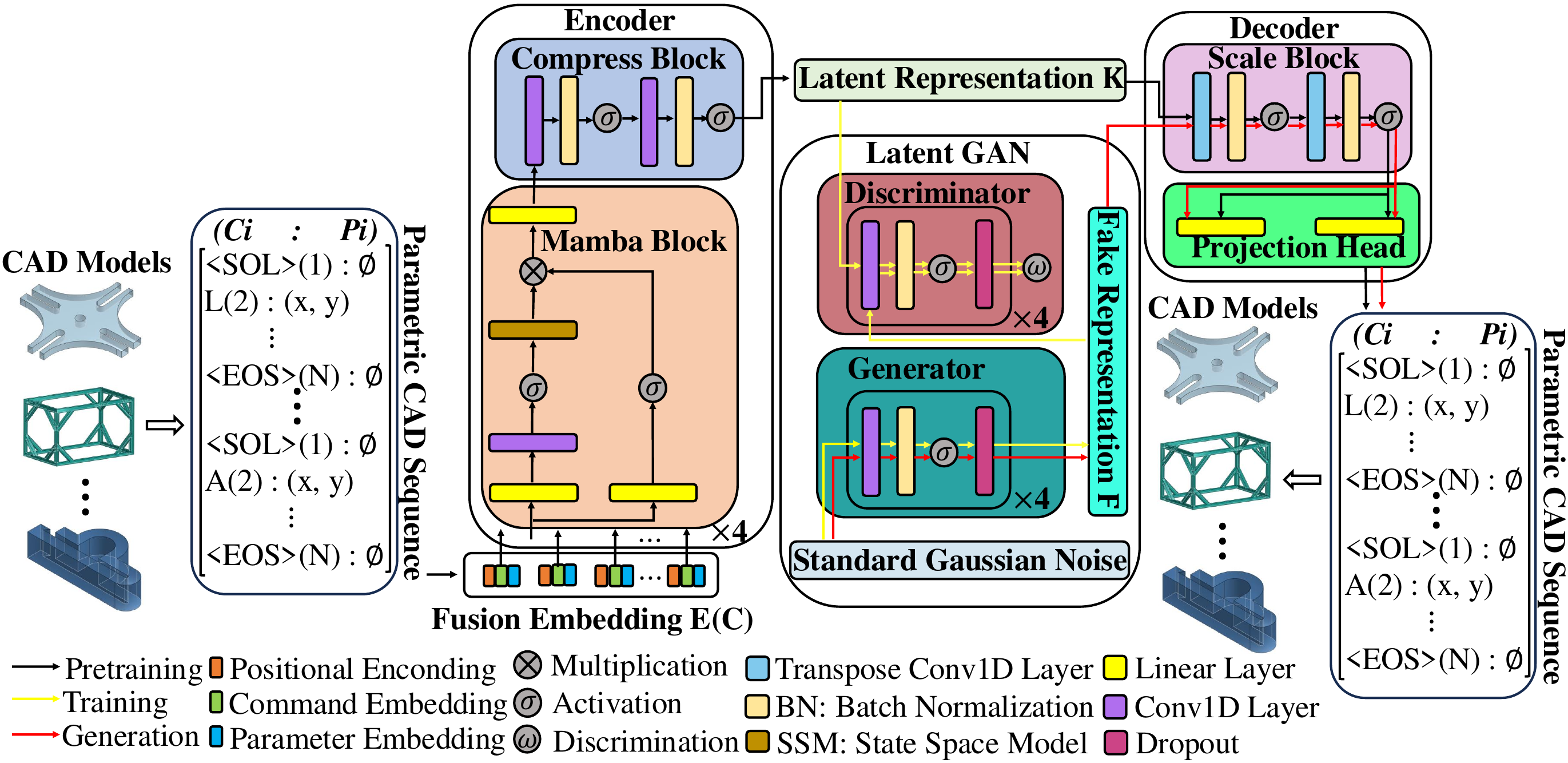} 
\caption{The overview of our Mamba-CAD architecture. The whole pipeline can be divided into three parts which marked in three different colored line-arrows. \textbf{(\textcolor{black}{Pre-training stage $\longrightarrow$})}: We leverage a task of CAD sequence reconstruction to pre-train the encoder and decoder of Mamba-CAD. Given a mini-batch of CAD-models, their corresponding parametric CAD sequences would be first discretized into friendly network representations (Sec. CAD sequence representation~\ref{sec2}). Next, these representations keep going forward Fusion Embedding, Encoder, and Decoder of Mamba-CAD (Sec. Mamba-CAD~\ref{sec3}) to recover predicted parametric CAD sequences. \textbf{(\textcolor{yellow}{Training stage $\longrightarrow$})}: When the pre-training stage is complete, the encoder parameters of Mamba-CAD would be frozen to produce the latent representation $\mathbb{K}$ that is further used to guide a 1-D Latent-GAN training. \textbf{(\textcolor{red}{Generation stage $\longrightarrow$})}: Once the 1-D Latent GAN is well-trained, it enables to randomly sample the noise from a standard gaussian distribution and further feed it into Generator to produce the fake latent representation $\mathbb{F}$, which would further undergo the decoder of Mamba-CAD to recover the parametric CAD sequence.}
\label{fig3}
\end{figure*}

In this section, we first introduce the rational logic of our designed Mamba-CAD and then present details of each component in Mamba-CAD. The whole architecture of Mamba-CAD can be overviewed in Figure~\ref{fig3}. Our objective is to generate parametric CAD sequences from randomly sampling the noise from a standard gaussian distribution.\\
\textbf{Preliminary.} The CAD models can be represented by parametric CAD sequences which are similar to discrete languages, inspiring us to leverage the sequential model to process CAD data. Mamba~\cite{r:8}, one of the discrete versions of recent State Space Model (SSM) shows the powerful capability of modeling on long sequences, which is very suitable for CAD models in the industry as they are relatively complex to require longer parametric CAD sequences to define. Hence, we make a step forward to explore the generative capability of Mamba on handling the complex CAD models requiring longer parametric CAD sequences. For a quick review, The origin of Mamba starts from SSM, which is conceptualized based on continuous systems with the ability to map a 1-D function or sequence $X(t)\in R \to Y(t)\in R$ via a hidden state $H(t)\in R^{N}$. Formally, the SSM model can be defined with Ordinary Differential Equation (ODE): $H'(t) = A\cdot H(t)+B\cdot X(t), Y(t)=\mathbb{C}\cdot H(t)$. Next, the continuous parameters A, B in this ODE would further be discretized to parameters $\mathbb{A}, \mathbb{B}$ via employing the Zero-Order Hold (ZOH) method with a timescale parameter $\Delta$, as defined with:
\begin{equation}
\mathbb{A}=exp(\Delta A), 
\label{eq1}
\end{equation}
\begin{equation}
\mathbb{B}=(\Delta A)^{-1}(exp(\Delta A)-I)\cdot\Delta B,
\label{eq2}
\end{equation}
\begin{equation}
H(t)= \mathbb{A}\cdot H(t-1)+\mathbb{B}\cdot X(t), 
\label{eq3}
\end{equation}
\begin{equation}
Y(t)=\mathbb{C}\cdot H(t),
\label{eq4}
\end{equation}
Based on the original SSM model, Mamba further introduces a Selective Scan Mechanism as its core SSM operator. For more details please refer to ~\cite{r:8}.
\\
\textbf{Embedding.} Given the parametric CAD sequence can be seen as a special discrete ``language", it inspires us to project CAD commands into an embedding space. As each CAD command is comprised by two parts: types of commands $c$ and their parameters $p$, we follow ~\cite{c:1} to separately map them into a two highly continuous space and then integrate them into a fusion embedding $E(C)$ as defined with:
\begin{equation}
E(C_{i}) = e_{c}^{(c_{i})} + e_{p}^{(p_{i})} + e_{pos}^{i},
\label{eq5}
\end{equation}
where $e_{c}^{(c_{i})}$ accounts for types of CAD commands $c_{i}$,  calculated through $e_{c}^{(c_{i})} = w_{c_{i}}\delta_{i}^c$. Here, $w_{c_{i}}\in R^{d_{E}\times{k}}$ is a learnable matrix. $\delta_{i}^c\in R^{k}$ denotes $c_{i}$ within the $k$ command types. $e_{p}^{(p_{i})}$ is the embedding of command parameters $p_{i}$, given by ${e}_{p}^{(p_{i})}=w_{p_{i}}^{a}f\left(w_{p_{i}}^{b} \delta_{i}^{p}\right)$. $f(*)$ flattens the matrix to a vector. Each CAD command is composed of 16 parameters and each of them can be quantized into an 8-bit integer. Including one additional dimension for those unused CAD commands, the integer is converted into a one-hot vector with dimension of $2^8+1=257$, thereby stacking these 16 one-hot vectors into a matrix $\delta_{i}^{p}\in R^{257\times16}$ . $w_{p_{i}}^{a}\in R^{d_{E}\times{16d_{E}}}$ and $w_{p_{i}}^{b}\in R^{d_{E}\times{257}}$ are learnable matrices. $e_{pos}^{i}$ is the positional encoding the same as in Transformer~\cite{r:4}.\\
\textbf{Encoder.} With the feed-forward strategy, a fusion embedding $E(C)$ as the input for encoder, which is composed of four Mamba Blocks, each with vocabulary size of 256 to match the dimension of a fusion embedding ($d_{E}=256$). After passing forward four Mamba Blocks, each $E(C)$ would be mapped into a vector $l\in R^{B\times256\times L}$, where the sequence length $L=128$ and batch size $B=32$ are set up practically. Considering the random generation as one of our designed tasks, it needs generate a fake latent representation from a standard gaussian noise and make it close enough to $l$. To make $l$ easy to learn during the generation stage, we further employ a Compress Block consisting of two 1-D Convolutional layers to reduce the feature dimension of $l$ from 256 to 64, namely the latent representation $\mathbb{K} \in R^{B\times\ 64\times L}$. According to Equation~\ref{eq4}, the feed-forward of encoder in Mamba-CAD can be factorized as:
\begin{equation}
\mathbb{K} = F[\mathbb{C}\cdot H(E(C_{i}))|S(6)],
\label{eq6}
\end{equation}
where $\mathbb{C}\in R^{N\times1}$ is a matrix projection and the function $F[*]$ denotes Compress Block. $S(6)$ is a selective mechanism of SSM model adopted in Mamba.\\
\textbf{Decoder.} In order to restore the scale of the latent representation $\mathbb{K}$, we first employ a Scale Block consisting of two 1-D Transpose-Convolutional layers with Batch Normalization~\cite{r:28} to turn it back. Next, we conduct two linear layers to decode types of commands $c_{i}^{*}$ and their parameters $p_{i}^{*}$ separately from $\mathbb{K}$, and make them gathered into a final predicted parametric CAD sequence $S^{*}=G[(c_{i}^{*}, p_{i}^{*})_{1}, (c_{i}^{*}, p_{i}^{*})_{2},\ldots,(c_{i}^{*}, p_{i}^{*})_{N}]$. We consider the whole Mamba-CAD as a feed-forward fashion and the output of our model can be formally defined with:
\begin{equation}
D(S^{*}|\mathbb{K}, \Psi) =\Pi_{j=1}^{N}G[D((c_{i}^{*}, p_{i}^{*})_{j}|\mathbb{K}, \Psi)],
\label{eq7}
\end{equation}
where $\Psi$ denotes the decoder parameters of Mamba-CAD.\\
\textbf{Pre-training.} As usual settings for pre-training an auto encoder-decoder network, we leverage a self-supervised task of CAD sequence reconstruction to train our Mamba-CAD. Specifically, the input is the parametric CAD sequence $S$ and would be fed into Mamba-CAD and undergo encoding and decoding to recover the predicted parametric CAD sequence $S^{*}$. The training objective is to make $S^{*}$ and $S$ close enough to each other, ensuring that Mamba-CAD learns well to model the parametric CAD sequence. This is important for learning the latent representation $\mathbb{K}$ that can transfer the knowledge of CAD models to subsequent random generation. In practical, we adopt the standard Cross-Entropy loss to minimize the distance from $S^{*}$ to $S$. Formally, the constraint of Mamba-CAD in pre-training is defined within one mini-batch $N_{m}$:
\begin{equation}
\mathcal{L}_{\mathrm{CE}} = -\sum_{j=1}^{N_{m}} (c_{i}, p_{i})_{j} \log (c_{i}^{*}, p_{i}^{*})_{j}.
\label{eq8}
\end{equation}
The pre-training is conducted on one NVIDIA RTX 3090 GPU with the batch size of 32 under 10 epochs in about 5 hours. The initial learning rate is set
to 0.001 with warm-up~\cite{c:22} and gradient clipping of 1.0 is applied in back-propagation.\\
\textbf{Training and Generation.} Once the Pre-training task is complete, we can leverage the latent representation $\mathbb{K}\in R^{B\times D\times L}$ produced by Compress Block as the well-learned knowledge of CAD models to further train a 1-D Latent GAN\cite{c:16} with 200,000 epochs and batch size of 256 in about 3 hours. To adapt to the main architecture of Mamba-CAD, we also use the same convolutional operations for Generator $\mathbb{G}$ and Discriminator $\mathbb{D}$. Practically, we use both four 1-D convolutional blocks consisting of Batch Normalization and Dropout (0.25)~\cite{r:27} operations for $\mathbb{G}$ and $\mathbb{D}$ (Figure~\ref{fig3}). Besides, we also adopt the same penalty mechanism as WGAN~\cite{r:29, c:1} to stabilize training. The training objective is to enable a gaussian noise to approach the latent representation $\mathbb{K}$ after passing forward Generator $\mathbb{G}$. When the network is well-trained, it enables to randomly sample the noise  from a standard gaussian distribution $\mu \sim \mathcal{N}(0,1)$ and feed it into Generator $\mathbb{G}$ to achieve a fake latent representation $\mathbb{F}$, which is highly similar to $\mathbb{K}$. Finally, $\mathbb{F}$ would pass forward the decoder of Mamba-CAD to recover the parametric CAD sequence, as factorized:
\begin{equation}
D(S^{*}|\mathbb{F}, \Psi) =\Pi_{j=1}^{N}G[D((c_{i}^{*}, p_{i}^{*})_{j}|\mathbb{F}, \Psi)],
\label{eq9}
\end{equation}
where the parameters $\Psi$ of decoder is frozen in the random generation stage. The pseudo code of Mamba-CAD can be found in the uploaded ``Technical Appendix".
\subsection{Dataset collection}\label{sec4}
Although several datasets of CAD models are publicly available, none of them can be directly used for training Mamba-CAD. Specifically, the ABC dataset~\cite{c:18} collected by~\cite{m:1} includes over 1 million CAD models defined with BRep representation, which lacks workflow constructions to make it difficult to be parameterized. DeepCAD~\cite{c:1} is a large-scale CAD dataset defined with CSG representation, consisting of 178,238 CAD models and their parametric CAD sequences. However, the maximize length of parametric CAD sequences within DeepCAD dataset is up to 60 and the average length of them is not long enough. Another Fusion360 Gallery~\cite{r:26} provides a sub reconstruction dataset consisting of about 7,000 CAD models as represented in parametric CAD sequences, which is also not enough to train a generative CAD model. To break through the limitation of parametric CAD sequences within the existing benchmark datasets, we further construct a new CAD dataset as represented in parametric CAD sequences. To this end, two steps are adopted to filter and collect CAD models. First, we again filter the DeepCAD dataset by discarding CAD models in which the length of their parametric CAD sequences is less than 10. Second, we further use the feature script from~\cite{m:1} to parse CAD models within ABC dataset and translate them into parametric CAD sequences by only retaining the CAD commands used in CSG representation. Similar to DeepCAD dataset, we further remove those CAD models which use commands beyond sketch and extrusion. Next, we set the length of parametric CAD sequences up to 128 to again filter these CAD models. Finally, total 77,078 CAD models are collected as represented in parametric CAD sequences, in which 61,662 CAD models for training, 7,707 models for validation and 7,709 models for testing. This is a CAD dataset with longer parametric CAD sequences than those in the DeepCAD dataset~\cite{c:1}. The parametric CAD commands contained in our dataset are listed in Figure~\ref{fig2}. We also statistically analyze the length of parametric CAD sequences in our new dataset and compare it with the benchmark DeepCAD dataset, including the average length and the length distribution, as shown in Table~\ref{table1}. It can be found that our dataset is significantly more complex than DeepCAD dataset in both the average length and length distribution of parametric CAD sequences.
\begin{table}[t]
\huge
\centering
\resizebox{.99\columnwidth}{!}{
\begin{tabular}{cccccccc}
\toprule
Dataset&TN&AL&[1-10]&[11-25]&[26-40]&[41-60]&[60-128]\\
\midrule
DeepCAD&178,238&15&44.58\%&38.73\%&11.38\%&5.31\%&-\\
Mamba-CAD&77,078&38&-&41.25\%&26.60\%&16.13\%&16.02\%\\
\bottomrule
\end{tabular}}
\caption{The statistic comparison of between DeepCAD dataset and Our Mamba-CAD dataset, including total number (TN), average length (AL) of parametric CAD sequences, and the length distribution.}
\label{table1}
\end{table}
\section{Experiments}
In this section, we mainly evaluate our Mamba-CAD in two aspects: (1) modeling on parametric CAD sequences; (2) random generation of parametric CAD sequences. Our goal here is to explore the capability of Mamba-CAD as a pure unimodal framework on handling the longer parametric CAD sequence, which is an open problem for previous efforts based on the Transformer network that only use parametric CAD sequences as inputs. 
\subsection{Evaluation Metrics and Comparison Methods}\label{EM}
\textbf{Evaluation metrics.} For evaluating the performance of modeling on parametric CAD sequences. Our model encodes each parametric CAD sequence $S$ into a latent space, and then decode them back to predicted CAD sequence $S^{*}$, $A_{c}$ and $A_{p}$ are used to evaluate the difference between $S$ and $S_{*}$ as defined below:
\begin{equation}
A_{c} = \frac{1}{N}\sum_{i=1}^{N}{\bigtriangledown [c_{t}^{i} = c_{t}^{i*}]}, 
\label{eq10}
\end{equation}
\begin{equation}
A_{p} = \frac{1}{T}\sum_{i=1}^{N}\sum_{j=1}^{K}{\bigtriangledown [c_{t}^{i} = c_{t}^{i*}]}\bigtriangledown[\left | c_{p}^{i,j} - c_{p}^{i,j*}\right | < \eta],
\label{eq11}
\end{equation}
where $\bigtriangledown[*]$ is a boolean function with scalar 0 or 1. $T$ is the total number of parameters in all correctly predicted commands. We set $\eta = 3$ in the experiments. The predicted sequence would probably be invalid to construct 3D shape. Therefore, we also calculate Invalid Ratio (IR) and Median Chamfer Distance (MCD) to better evaluate its rationality of shape construction. For its capability on random generation, we adopt three frequently used 3D metrics of Coverage (COV), Jensen-Shannon Divergence (JSD) and Minimum matching distance (MMD). We also report metrics of ``Unique" and ``Novel". The ``Unique" score is the percentage of data appears once in the generated set and the ``Novel" score is the percentage of the generated data does not appear in the training set. To evaluate Mamba-CAD in CAD sequence modeling and generation comprehensively, we further calculate the average length of generated or recovered parametric CAD sequences, which are valid to construct into final 3D shapes. Considering the STEP format is widely used in CAD tools (e.g., AutoCAD, SolidWorks), we also give a metric named Step Ratio (SR), which indicates the proportion of the parametric CAD sequences that can be successfully converted to STEP format.
\begin{figure}[t]
\centering
\includegraphics[width=.99\columnwidth]{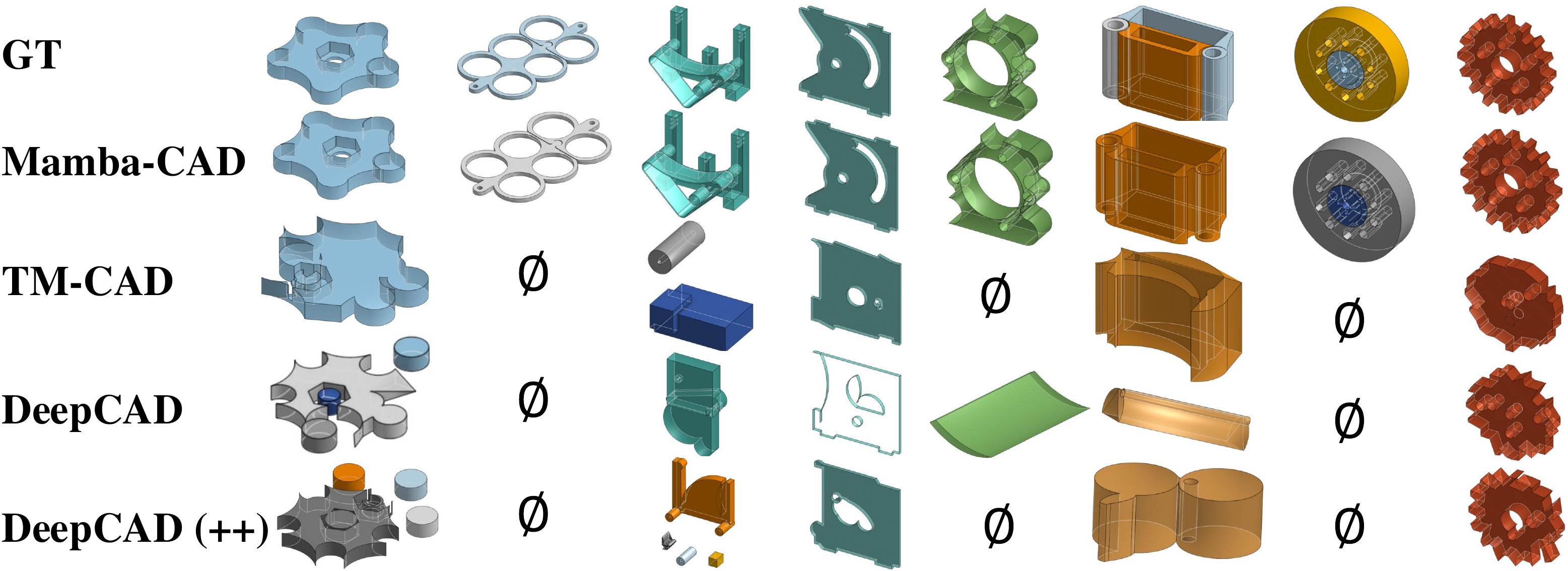} 
\caption{Visual illustrations of CAD sequence reconstruction. GT represents Ground Truth. $\emptyset$ denotes the CAD sequence is invalid, leading to the failure of constructing a 3D shape. Please enlarge 200\% for a clear view.}
\label{fig4}
\end{figure}\\
\textbf{Comparison methods.} Our primary goal is to explore the modeling capabilities of Mamba-CAD for parametric CAD sequences, rather than to prove the effectiveness of improving the modeling capability via adding additional modalities. Hence, the discussion here is based on unimodal approaches, modeling parametric CAD sequences. To make fair comparisons, all comparable models are re-trained on the Mamba-CAD dataset. In practical, we compare Mamba-CAD with following methods under several tasks to justify the effectiveness of Mamba-CAD in handling the longer parametric CAD sequence. \textbf{(1) DeepCAD}~\cite{c:1} is an auto-encoder framework based on Transformer network, which encodes parametric CAD sequences into a latent space and then decodes them back to construct 3D shapes. \textbf{(2) DeepCAD (++)} is an ultra version of DeepCAD, which is trained longer and composed of more Transformer layers and the larger dimension compared to DeepCAD. \textbf{(3) TM-CAD} is our another simple attempt at consolidating Transformer and Mamba into a fusion framework to process parametric CAD sequences. Practically, we leverage four Mamba Blocks to bridge the encoder and decoder of the framework based on Transformer network, where the settings of encoder and decoder are the same as those in DeepCAD. \textbf{(4) SkexGen}~\cite{c:2} is also a unimodal Transformer-based network but processes the ``sketch" and ``extrude" operations separately. \textbf{(5) HNC-CAD}~\cite{c:23} is a current state-of-the-art unimodal generative CAD model enabled by a two-stage cascaded auto-regressive Transformer-based network. Similar to Skexgen, HNC-CAD leverages a code tree to decompose one CAD model in there parts including ``loop", ``profile", and ``solid". Among these methods, both Skexgen and HNC-CAD decompose each CAD model into a specific representation, which is the core to fit their proposed frameworks. Hence, we follow the same pre-processing strategy as adopted in Skexgen and HNC-CAD to decompose CAD models within our Mamba-CAD dataset to ensure the compatibility with their frameworks training.
\begin{table}[t]
\huge
\centering
\resizebox{.99\columnwidth}{!}{
\begin{tabular}{lccccccc}
\toprule
Method&$A_{c}\uparrow$&$A_{p}\uparrow$&MCD$\downarrow$&IR$\downarrow$&AL$\uparrow$&L$\geq$60$\uparrow$&SR$\uparrow$\\
\midrule
DeepCAD&96.49&87.38&2.97&14.83\%&32.52&51.82\%&86.58\%\\
DeepCAD (++)&96.92&87.68&2.81&14.05\%&32.49&50.85\%&87.49\%\\
TM-CAD&95.73&85.48&4.82&15.85\%&31.21&50.31\%&85.61\%\\
\textbf{Mamba-CAD}&\textbf{99.96}&\textbf{99.93}&\textbf{0.64}&\textbf{5.41\%}&\textbf{38.91}&\textbf{85.18\%}&\textbf{95.43\%}\\
\bottomrule
\end{tabular}}
\caption{The comparison of Mamba-CAD and variants in CAD sequence reconstruction. MCD is multiplied by $10^{3}$. $A_{c}$, $A_{p}$, and IR are multiplied by 100\%. L$\geq$60: the ratio of valid reconstructed CAD sequences (Length$\geq$60) in total CAD sequences (Length$\geq$60) within test set.}
\label{table2}
\end{table}
\subsection{Modeling on parametric CAD sequences}\label{PR}
\textbf{CAD sequence reconstruction.} The ability to model the parametric CAD sequences is important for generative models, as it reflects whether the generative model can effectively represent the target distribution. Hence, we leverage a pre-training task of CAD sequence reconstruction to observe the performance of Mamba-CAD and comparable methods. Specifically, when the pre-training task is completed, 7,709 CAD models in test set would go forward networks to achieve reconstructed parametric CAD sequences. From Table~\ref{table2}, we can see that our Mamba-CAD surpasses other comparable methods on all metrics, especially in $A_{p}$, MCD, IR, SR, and L$\geq$60. This demonstrates that Mamba-CAD has superior modeling capabilities for long parametric CAD sequences compared to Transformer-based networks. Figure~\ref{fig4} gives more visual illustrations of the recovered CAD models from reconstructed parametric CAD sequences. As shown in Figure~\ref{fig4}, the CAD models recovered by Mamba-CAD is closer to the ground truth than other comparable methods, which again demonstrates the effectiveness of Mamba-CAD in handling longer parametric CAD sequences.
\begin{table}[t]
\huge
\centering
\resizebox{.95\columnwidth}{!}{
\begin{tabular}{lccccccc}
\toprule
Method&$A_{c}\uparrow$&$A_{p}\uparrow$&MCD$\downarrow$&IR$\downarrow$&AL$\uparrow$&L$\geq$60$\uparrow$&SR$\uparrow$\\
\midrule
DeepCAD&83.75&69.77&79.08&42.36\%&28.72&19.11\%&59.19\%\\
DeepCAD (++)&85.95&71.27&69.68&40.19\%&29.78&20.74\%&61.12\%\\
TM-CAD&80.75&67.71&82.34&44.34\%&28.26&17.79\%&57.23\%\\
HNC-CAD&-&-&-&-&30.51&42.89\%&74.36\%\\
\textbf{Mamba-CAD}&\textbf{99.85}&\textbf{98.91}&\textbf{1.03}&\textbf{8.03\%}&\textbf{35.43}&\textbf{79.46\%}&\textbf{92.84\%}\\
\bottomrule
\end{tabular}}
\caption{Quantitative results of CAD sequence completion.}
\label{table3}
\end{table}
\begin{table}[t]
\huge
\centering
\resizebox{.75\columnwidth}{!}{
\begin{tabular}{lccccc}
\toprule
Method&$A_{c}\uparrow$&$A_{p}\uparrow$&MCD$\downarrow$&IR$\downarrow$&SR$\uparrow$\\
\midrule
DeepCAD&95.23&85.58&10.97&10.78\%&90.34\%\\
DeepCAD (++)&96.79&87.72&8.93&9.35\%&91.22\%\\
TM-CAD&90.94&83.85&12.46&12.37\%&87.61\%\\
\textbf{Mamba-CAD}&\textbf{99.21}&\textbf{98.21}&\textbf{1.31}&\textbf{3.72\%}&\textbf{96.51\%}\\
\bottomrule
\end{tabular}}
\caption{The quantitative results of generalization ability in Fusion360 reconstruction dataset.}
\label{table4}
\end{table}
\begin{figure}[t]
\centering
\includegraphics[width=.9\columnwidth]{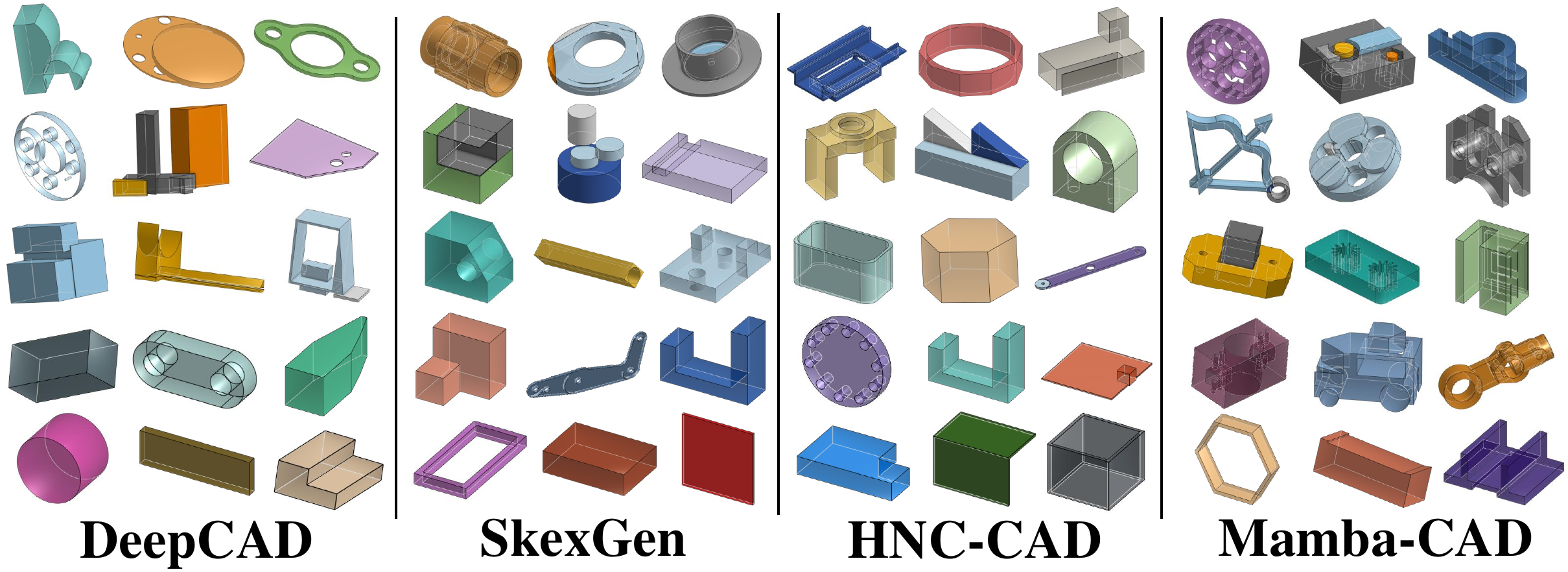} 
\caption{Visualizations of random generation of parametric CAD sequences.}
\label{fig5}
\end{figure}
\begin{table}[t]
\centering
\huge
\resizebox{.95\columnwidth}{!}{
\begin{tabular}{lccccccc}
\toprule
Method&MMD$\downarrow$&JSD$\downarrow$&COV$\uparrow$&Unique$\uparrow$&Novel$\uparrow$&AL$\uparrow$&SR$\uparrow$\\
\midrule
DeepCAD&3.51&8.93&63.77\%&83.41&51.94&27.78&62.76\%\\
SkexGen&2.96&7.39&65.89\%&93.33&60.47&29.18&63.49\%\\
HNC-CAD&2.65&6.07&70.43\%&92.18&55.32&31.36&68.84\%\\
\textbf{Mamba-CAD}&\textbf{1.94}&\textbf{4.28}&\textbf{75.61\%}&\textbf{94.73}&\textbf{63.25}&\textbf{36.75}&\textbf{74.13\%}\\
\bottomrule
\end{tabular}}
\caption{The quantitative results of random generation of parametric CAD sequences. MMD, JSD, Unique, and Novel are all multiplied by $10^{2}$.}
\label{table5}
\end{table}
\begin{table}[t]
\huge
\centering
\resizebox{.95\columnwidth}{!}{
\begin{tabular}{lccccccc}
\toprule
Method&$A_{c}\uparrow$&$A_{p}\uparrow$&MCD$\downarrow$&IR$\downarrow$&AL$\uparrow$&L$\geq$60$\uparrow$&SR$\uparrow$\\
\midrule
A&93.67&83.41&7.89&20.79\%&26.93&43.57\%&79.36\%\\
B&94.23&83.95&5.67&18.04\%&30.48&50.26\%&82.77\%\\
C&99.91&99.57&0.69&6.29\%&38.16&84.53\%&93.91\%\\
\textbf{Mamba-CAD}&\textbf{99.96}&\textbf{99.93}&\textbf{0.64}&\textbf{5.41\%}&\textbf{38.91}&\textbf{85.18\%}&\textbf{95.43\%}\\
\bottomrule
\end{tabular}}
\caption{Ablation studies of Mamba-CAD.}
\label{table6}
\end{table}\\
\textbf{CAD sequence completion.} To further evaluate the effectiveness of Mamba-CAD in learning CAD representations in the latent space, we randomly mask 40\% of each parametric CAD sequence with scalar 0 in the test set and make it go forward networks to achieve complete parametric CAD sequences. Table~\ref{table3} indicates that even with incomplete parametric CAD sequences as inputs, Mamba-CAD still achieves precise reconstruction results, again validating the effectiveness of our method in processing longer parametric CAD sequences. Besides, The advantages of Mamba-CAD are amplified compared to other models (e.g., the metrics of IR and MCD widen the gap). This completion task again demonstrates the superior ability of Mamba-CAD in context modeling for longer parametric CAD sequences, enabling it to accurately fill in the missing parts.\\
\textbf{Generalization ability.}
To further evaluate the generalization capability of Mamba-CAD, we make it first well-trained on the Mamba-CAD dataset and then directly test it on the Fusion360 reconstruction dataset that consists of about 7, 000 CAD models as represented in parametric CAD sequences. Although the CAD models in this dataset is simpler (e.g., CAD sequence length $\leq$60) than those in Mamba-CAD dataset, which narrows the gap of Step Ratio between Mamba-CAD and other Transformer-based networks, Mamba-CAD still achieves the best scores in all metrics. The quantitative results can be found in Table~\ref{table4}.
\subsection{Random generation of parametric CAD sequences}\label{RG}
When our framework is well-trained, we randomly sample total 10,000 noise vectors from a standard gaussian distribution as fake representations of CAD models and feed them into the decoder of Mamba-CAD to generate 10,000 parametric CAD sequences. For other comparable methods 10,000 samples are random generated to compare with Mamba-CAD. To evaluate Mamba-CAD and comparable methods in random generation quantitatively, we further convert generated CAD models into point clouds and calculate the related metrics. The results can be found in Table~\ref{table5}. Figure~\ref{fig5} shows that the created CAD models from Mamba-CAD are more complex than those created by Transformer-based networks. It again proves Mamba-CAD outperforms other comparable methods in processing longer parametric CAD sequences to achieve  more complex CAD models. Besides, from the showcases at the bottom of Figure~\ref{fig5}, it shows Mamba-CAD also has the ability to generate simple CAD models with short parametric CAD sequences.
\subsection{Ablation study}\label{ab}
To better understanding the mechanism of Mamba-CAD, we design three more ablation studies based on the CAD sequence reconstruction task: \emph{(A) replace Mamba block with Transformer block}, \emph{(B) replace CNN layers with MLPs in the Compress Block and Scale Block}, \emph{(C) add bottleneck layer between Encoder and Decoder}. From Table~\ref{table6}, it shows Mamba-CAD surpasses other comparable versions, which indicates the design components in Mamba-CAD are effective and reasonable. Especially compared to (\emph{A}), Mamba block brings significant improvement to the CAD sequence reconstruction task, which again proves its capability of processing longer parametric CAD sequences.
\section{Discussion and Conclusion}\label{ds}
Although Mamba-CAD shows a significant capability of modeling on parametric CAD sequences, there are still some limitations in our method. First, the types of commonly used CAD commands in practical designs are more than CAD commands discussed in our paper. For example, one specific 
curve of Bézier has not been parameterized and defined in our dataset, which imposes limitations on the topological structures of CAD models generated by our approach. Second, Mamba-CAD cannot parse CAD commands such as ``fillet" and ``chamfer" from BRep representation. To combine CAD commands from CSG and BRep representation for generative CAD modeling would be an innovative touch in the future work. Third, Mamba-CAD mainly focuses on the unconditioned generative modeling of longer parametric CAD sequence. However, other modalities such as texts, images, and point clouds can be integrated as an additional condition to further make the generation controllable. This is also an interesting task left for the future work.

In summary, we introduces Mamba-CAD, a deep generative model based on a Mamba backbone for CAD designs, which can handle a longer parametric CAD sequence to construct complex 3D shapes of CAD models. To achieve this, we also create a new CAD dataset consisting of 77, 078 CAD models represented in parametric CAD sequences. The average length of CAD sequences in the new dataset is longer than that in benchmark datasets of the same type.
\bibliography{aaai25}
\end{document}